# Exploiting Digital Surface Models for Inferring Super-Resolution for Remotely Sensed Images

Savvas Karatsiolis, Chirag Padubidri and Andreas Kamilaris

*Abstract*— Despite the plethora of successful Super-Resolution Reconstruction (SRR) models applied to natural images, their application to remote sensing imagery tends to produce poor results. Remote sensing imagery is often more complicated than natural images and has its peculiarities such as being of lower resolution, it contains noise, and often depicting large textured surfaces. As a result, applying non-specialized SRR models like the Enhanced Super Resolution Generative Adversarial Network (ESRGAN) on remote sensing imagery results in artifacts and poor reconstructions. To address these problems, we propose a novel strategy for enabling an SRR model to output realistic remote sensing images: instead of relying on feature-space similarities as a perceptual loss, the model considers pixel-level information inferred from the normalized Digital Surface Model (nDSM) of the image. This allows the application of better-informed updates during the training of the model which sources from a task (elevation map inference) that is closely related to remote sensing. Nonetheless, the nDSM auxiliary information is not required during production i.e., the model infers a super-resolution image without additional data. We assess our model on two remotely sensed datasets of different spatial resolutions that also contain the DSMs of the images: the DFC2018 dataset and the dataset containing the national LiDAR fly-by of Luxembourg. We compare our model with ESRGAN and we show that it achieves better performance and does not introduce any artifacts in the results. In particular, the results for the high-resolution DFC2018 dataset are realistic and almost indistinguishable from the ground truth images.

## I. Introduction

High-quality aerial photography and satellite imagery facilitate the development of interesting remote sensing applications for large-scale monitoring and earth observation, including land monitoring, urban planning, and surveillance. However, the severe weakness of remotely sensed imagery, in general, is its low spatial resolution, i.e., the detail level is insufficient for detecting certain objects of interest like tree types, solar panels on rooftops and cars. Remotely sensed imagery usually has low spatial resolution due to the cost and time required to collect high-quality/low-noise images and the vulnerability of such images to environmental variations during acquisition like atmospheric and light variations. Excessive costs may diminish the advantages of using high-quality imagery in remote sensing applications. A common alternative to address this limitation is the use of low-quality (low spatial resolution/high noise) images to reconstruct scene information as much as possible and then perform inference based on the information-enriched data. This strategy maintains the lower image acquisition cost and improves the quality of the final output.

Traditional upsampling methods such as nearest-neighbor and bi-cubic interpolation [1] rely on surrounding pixels to add a small amount of information to an image and tend to produce blurry and distorted results mainly because they fail to recover high-frequency information. Inevitably, demanding applications like small-object detection tasks do not generally benefit much from image interpolation by methods that rely on neighboring pixels to add some level of detail to the image. Some early attempts to produce better results than the traditional interpolation methods involved the learning of degradation models [2] and the feature matching of low/high-resolution patches to facilitate the recovery of high-resolution (HR) images [3]–[5]. Slightly more sophisticated methods built sparse representations that comprised of a dictionary used to reconstruct the HR counterpart of a low-resolution (LR) patch [6], [7]. While image upsampling using sparse representations tends to slightly improve the recovery of high-frequency information, it is a very computationally intensive technique [7]. The limited performance of these approaches in effectively converting an LR image to its realistic HR counterpart originates from their inability to learn.

One of the reasons that Deep Learning (DL) and Convolutional Neural Networks (CNNs) have become extremely popular is their ability to supply end-to-end models that perform inference based on raw data without relying on hand-engineered features or extensive incorporation of task-related knowledge into the model. These characteristics positioned DL as the mainstream approach nowadays to solving challenging remote sensing tasks. As such, DL is widely used for tackling the Super-Resolution Reconstruction (SRR) task i.e., converting a single LR image to an HR one. The output of the SRR task is called a Super Resolution (SR) image and its goal is to learn how to produce SR images from LR images that are indistinguishable from the ground truth i.e., the HR images. Furthermore, the development of efficient SSR models will

Corresponding author: Savvas Karatsiolis.
Savvas Karatsiolis is with the CYENS Center of Excellence, Nicosia, Cyprus (e-mail: s.karatsiolis@cyens.org.cy).
Chirag Padubidri is with the CYENS Center of Excellence, Nicosia, Cyprus (e-mail: c.padubidri@cyens.org.cy).

Andreas Kamilaris is with the department of Computer Science, University of Twente, Enschede, The Netherlands (e-mail: a.kamilaris@utwente.nl) and CYENS Center of Excellence, Nicosia, Cyprus (email: a.kamilaris@cyens.org.cy)

greatly benefit DL models performing a plethora of remote sensing tasks since high-quality imagery is especially beneficial for DL models [8]–[11]. With the ever-increasing usage of DL methodologies for developing remote sensing applications [12]–[16], training and inferring on HR images greatly increases the chances of obtaining good results on notoriously difficult tasks.

## II. RELATED WORK

The first attempts of using DL for the SSR task used pixel loss between the SR output and the HR image (ground-truth). Pixel losses are straightforward to implement. Specifically, minimizing the mean squared error (MSE) conveniently maximizes the peak-signal-to-noise ratio (PSNR), which is a commonly used measure for evaluating SRR models. However, PSNR is not a good measure of perceptual similarity because it fails to capture perceptually relevant differences [17]. In particular, the textual detail level is not reflected in the magnitude of the measured PSNR. Pixel losses tend to produce overly smoothed outputs that constitute candidate HR reconstructions: the model calculates a statistical average of the plausible HR reconstructions introduced to it during training. Super-Resolution CNN (SRCNN) [18] was one of the early attempts that used a DL model trained on a pixel loss for the SRR task. Many following attempts experimented with various advanced architectural features in the DL model to mitigate the effects of pixel losses. Kim et al. [19] applied residual learning [20] into a very deep CNN, Zhang et al. [21] applied deep residual channel attention mechanisms and Lai et al. [22] proposed the Laplacian super-resolution network (LapSRN), which supported high up-sampling factors with the use of residual skip connections. Despite the extensive focus on identifying novel architectural features that improve SRR models, the gap between the quality of the HR images and the SR outputs remained. To overcome the limitations created by applying a pixel loss between the ground-truth and the SR image, Johnson et al. [23] introduced a perceptual loss to measure semantic similarity between the two images. They specifically used a VGG-16 [24] model trained on ImageNet [25] and minimized the Euclidean distance between the features of the HR images and the features of the SR images (i.e., a perceptual loss). They showed that this strategy allowed the model to reconstruct fine details and edges. These results are in line with Mahendran and Vedaldi [26] who also showed that matching the features of higher layers in the pre-trained model preserves the image content and the spatial structure of an image. Johnson et al. [23] trained two SRR models: one that did not use any pixel loss during training and relied solely on the perceptual loss and one that only used a pixel loss. The outputs produced by the two models confirmed that while the perceptual loss is better at reconstructing minute details and producing visually appealing results, the pixel loss gives much fewer artifacts mainly because of its smoothing effect on the pixel values. This result suggests that both losses are useful for the SRR task.

Further, Generative adversarial networks (GANs) [27] are highly effective generative models for producing realistic images. The GAN learns a mapping from one manifold to another via an adversarial game between a generator model and a real/generated image discriminator model. GAN's ability to produce sharp images by learning the actual data distribution [27] suggests that the adversarial loss might be a good fit for the SRR task. Indeed, Ledig et al. [17] proposed a GAN-based model for the super-resolution task (SRGAN), combining three losses: a content loss (MSE pixel loss), a perceptual loss (VGG feature matching like in [23]), and an adversarial loss that encourages the network to favor solutions that reside on the manifold of natural images. Wang et al. [28] proposed some improvements to SRGAN including a) the implementation of Residual-in-Residual Dense Blocks (RRDBs), which constitute an extension to densely connected networks [29], b) the use of relativistic adversarial loss [30] which stabilizes the GAN's training and improves its performance and c) the application of the perceptual loss before the activations of the VGG layers. Wang et al. called their improved model Enhanced Super-Resolution GAN (ESRGAN). While the ESRGAN's performance on natural images is quite impressive, it tends to create artifacts in remotely sensed imagery [31]. This may emanate from the complexity and variability of the scenes depicted in remote sensing [31] or from the images' lower spatial resolution and the higher noise they usually exhibit. Furthermore, a huge portion of remotely sensed images often includes textured surfaces, in contrast to the images contained in the ImageNet dataset that have more high-frequency components spread throughout the image area.

These peculiarities of remotely sensed images are better managed by models that are oriented to work with such data. In this direction, Gong et al. [31] proposed the Enlighten-GAN model that uses a self-supervised hierarchical perceptual loss. Liu et al. [32] exploited the salient maps of images to learn additional structure priors and to make the model focus more on the salient objects. Huan et al. [33] proposed a multi-scale residual network with hierarchical feature fusion and multi-scale dilation residual blocks. Courtrai et al. [34] used a cycle-GAN [35] to convert LR images to HR images as well as HR images to LR images, which is a process that seems to help the model learn the mapping between the two domains. Courtrai et al. also integrated a YOLOv3 [36] model into their architecture to conduct small object detection. The integrated object detection model, together with the cycle-GAN, train the generator synergically. Despite small object detection being the model's main aim, the generator produces upsampled images to facilitate the task.

Summing up, previous works on the SRR task for remote sensing imagery focus either on the architecture of the model or on small training procedure differentiations that potentially improve the results to a certain extent. In this paper, the authors exploit the best practices derived from state-of-the-art experimentation up to date, suggesting partly keeping the training principles (content and adversarial losses) of the highly successful ESRGAN while modifying the way the perceptual loss is conceived in the context of super-resolution in general and in remote sensing specifically. The key idea of the proposed approach lies in the observation that single-image super-resolution is an ill-posed problem in the sense that for any LR image exist numerous HR images that could correspond to it [23], [31], [32]. Thus, for any successful model to achieve superior performance, it must derive significant pixel-level knowledge during the training. Up to date, most promising models use a perceptual loss that is based on feature matching, i.e., matching the similarity of two images in feature space [17], [23], [28]. Alternatively, the authors propose the replacement of the perceptual loss with a pixel-level loss which is more appropriate for SRR models operating on remotely sensed imagery. Specifically, the authors exploit the normalized Digital Surface Model (nDSM), defined as the difference between the Digital Surface Model (DSM) and the Digital Elevation Model (DEM), i.e., nDSM=DSM-DEM. The nDSM holds a great amount of pixel-level information that can restrain the model's flexibility in outputting a statistical average of viable solutions: a candidate solution must have the same nDSM as the ground-truth HR image. The main difference between the proposed approach and a feature matching perceptual loss is that the nDSM contains most of the spatial relations within an image while feature space similarities may be misleading: semantically unrelated images can have a similar subset of features. Johnson et al. [23] also note this while discussing which VGG layers (lower-level or higher-level) to choose when constructing the perceptual loss. The results of this paper suggest that the gradients flowing from the nDSM back to the SRR model during training improve the quality of the latter. Some task-specific training techniques have also been applied, which stabilize training and improve the results.

## III. USING DSMs TO APLLY SUPER RESOLUTION

As mentioned in related work, the ESRGAN model's performance in upsampling natural images is impressive in part due to the perceptual loss used during training. The perceptual loss of the ESRGAN is calculated using a second model pre-trained on a second task relevant to the primary task of interest e.g., a classification task conducted via a VGG model. Regarding the ESRGAN's training, the VGG model used for calculating the perceptual loss was trained on the *1000* classes of the ImageNet dataset. This substantial number of classes, in combination with the millions of images contained in the dataset and the effectiveness of the VGG model, facilitated the training of the SRR task. However, remote sensing imagery differs from the images contained in the ImageNet dataset in several aspects: they have lower spatial resolution and level of detail, they have higher noise and they depict larger textured surfaces instead of individual objects dominating the image. Thus, a pre-trained VGG model on the ImageNet dataset might not be the best choice for training an SRR model that takes as input aerial photography or satellite imagery. Even if an ESRGAN-like model is trained on a remote sensing task from scratch, using its learned features for building a perceptual loss, it will most probably not be exposed to hundreds of classes or have access to millions of high-resolution images. Such limitations are quite common when dealing with remote sensing tasks. Furthermore, both large textured surfaces and image variability in remote sensing imagery tend to reduce the effectiveness of feature space similarity metrics. In the following subsections, we describe the loss functions used in the proposed SRR model.

*A. The nDSM-based loss*

The authors address the problems imposed on the SRR task of remote sensing imagery by applying a pixel-level loss based on information that is closely related to the domain, harnessing the nDSM. An nDSM inferring model captures the spatial relations in remotely sensed images to infer the heights of the depicted objects. Interestingly, neural networks (NN) predicting depth from single images also use object interactions, like shadows, to identify objects in the scene [37]. To infer the nDSMs, we use a model developed in previous work [38] that converts single RGB images to nDSMs and we pre-train it with the data used for the SRR task. The model inferring the nDSMs from RGB images uses a U-Net architecture [39] to compress an image into smaller representations that the model then decodes to form the elevation map. For a detailed model architecture and details regarding its training, we kindly refer the readers to [38]. Figure 1 shows examples of inferring the elevation map of an aerial image via the nDSM model.

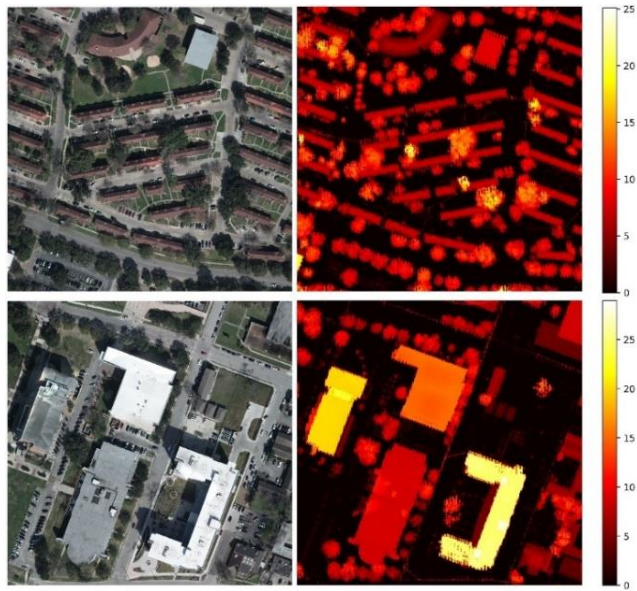

**Fig. 1.** The nDSM model [38] infers the height map of the objects depicted in an aerial image. The RGB aerial images are shown on the left and the predicted elevation heat maps are shown on the right. The color bars indicate the color-coded height in meters.

The pre-trained nDSM provides the means for defining a loss that closely relates to the domain data. Besides the nDSM-based loss, our proposed SRR methodology also uses a content loss (pixel-loss) and an adversarial loss. The content loss forces the model to output images that maintain the content of the LR image while the adversarial loss drives the model to infer images that are sharper and more realistic.

Ideally, an SR image calculated from its LR counterpart should result in the same nDSM as the ground-truth HR image corresponding to the LR image. The closer an SR image is to the ground-truth HR image, the closer their inferred nDSMs should be. This is reflected in the following loss:

$$L_{nDSM} = \|f_{nDSM}(f_{SR}(x_{LR})) - f_{nDSM}(x_{HR})\|_2 \quad (1)$$

where $f_{nDSM}(.)$ is the nDSM-inferring model, $f_{SR}(.)$ is the SRR model, $x_{LR}$ is the LR image and $x_{HR}$ is the HR image. During training, the LR image is used as the input of the SRR model and its SR reconstruction is passed through the pre-trained nDSM model. Then, the ground-truth HR image is also passed through the nDSM model, and the Euclidean distance of the inferred height maps is calculated. During production, the nDSM model is no longer required, since its sole purpose is to facilitate the model training by forcing the parameters' update operation to favor weights that output SR images that are similar to the HR images.

*B. Adversarial loss*

In addition to the content loss and the nDSM loss, we also use an adversarial loss to bias the model toward images that reside on the HR images' manifold. We adopt the original GAN methodology proposed by SRGAN [17] and not the relativistic GAN suggested in ESRGAN [28] because we did not observe any improvements in our results when the latter was used. However, we noticed that the proposed nDSM loss stabilizes the training and facilitates learning, which makes the use of a vanilla GAN sufficient. The adversarial loss is defined as

$$L_{adversarial} = \sum_n -\log f_{DGAN}(f_{SR}(x_{LR})) \quad (2)$$

with $f_{DGAN}$ being the discriminator of the GAN, $f_{SR}$ is the SRR model and $x_{LR}$ is the LR image. We use this formulation instead of minimizing $\log(1 - f_{DGAN}(f_{SR}(x_{LR})))$ to avoid the saturating gradient issue [27]. The GAN discriminator is trained on predicting whether input images source from the HR images' distribution or the SR images' distribution.

*C. The proposed super-resolution model architecture*

ESRGAN employs the basic architecture of SRGAN [17], incorporating modifications such as the removal of batch normalization [40] everywhere in the model and the use of RRDBs as the basic block of the model. The specific architecture performs most computations in the LR feature space and uses up-scaling units located near the output which increases the resolution of the feature maps calculated by the RRDBs. We apply some further modifications to the ESRGAN architecture that enhance its performance on remote sensing imagery:

1) Each sub-pixel up-sampling (x2) layer is followed by two convolutional layers with Parametric Rectified Linear Units (PReLUs) [41].
2) After the up-sampling units, we use two additional convolutional layers with PReLU activations.
3) The output convolutional layer applies a hyperbolic tangent activation function followed by an operation that converts the resulting values in the range [0,1]. Specifically, the rescaling operation applies the function $0.5 \times (x - 1) + 1$ to an input $x$.

Figure 2 shows the proposed modified architecture for the super-resolution network. The modifications made to the original ESRGAN architecture are noted in Figure 2. The model implements several blocks, each consisting of three RRDBs that contain residual nodes and dense connections. Each residual node applies a scaling parameter $\beta$ to the output of each RRDB before adding it to the residual path. A similar scaling is applied at the output of each RRDB and specifically at the residual node that merges the input of the block with its output. Residual scaling prevents instability during the training and allows for smoother updates [28].

*D. Content loss*

Most SRR models use either the Mean Squared Error (MSE) or Mean Absolute Error (MAE) to implement the content loss. These error functions tend to be a good fit for close-range photography, but this might not be the case for remotely sensed imagery. As mentioned before, aerial images usually have low spatial resolution and are noisy. Large texture surfaces of a wide variety make the SRR task even harder. Images of rocky areas, random soil formations, dumping fields with randomly disposed waste and varied objects' orientations make it extremely hard for a DL model to learn the data distribution. Furthermore, trees with entangled branches and an infinite number of leaves configurations render the SRR task on remote sensing imagery extremely hard even for state-of-the-art models like the ESRGAN. In particular, soil and leaves' configurations are very complex and thus very hard to model in the super-resolution context. MSE penalizes large prediction errors which makes MAE more suitable when the dataset contains several outliers. In the case of aerial imagery-based SRR, a wise strategy is to avoid high penalization on the reconstruction error of entities whose distribution is a priori difficult to learn (e.g., soil and trees) and to penalize large errors on easier-to-learn reconstructions like cars and houses. Thus, we propose the use of the Huber loss [42] instead of MAE or MSE because it applies either of the two losses, depending on the error magnitude. The proposed content loss is a Huber function with a transition point $\epsilon$ and $a = f_{SR}(x_{LR}) - x_{HR}$ defined as:

$$L_{content} = \begin{cases} \frac{1}{2}(a)^2 & if\ |a| \leq \epsilon \\ \epsilon(|a|) - \frac{1}{2}\epsilon & otherwise \end{cases} \quad (3)$$

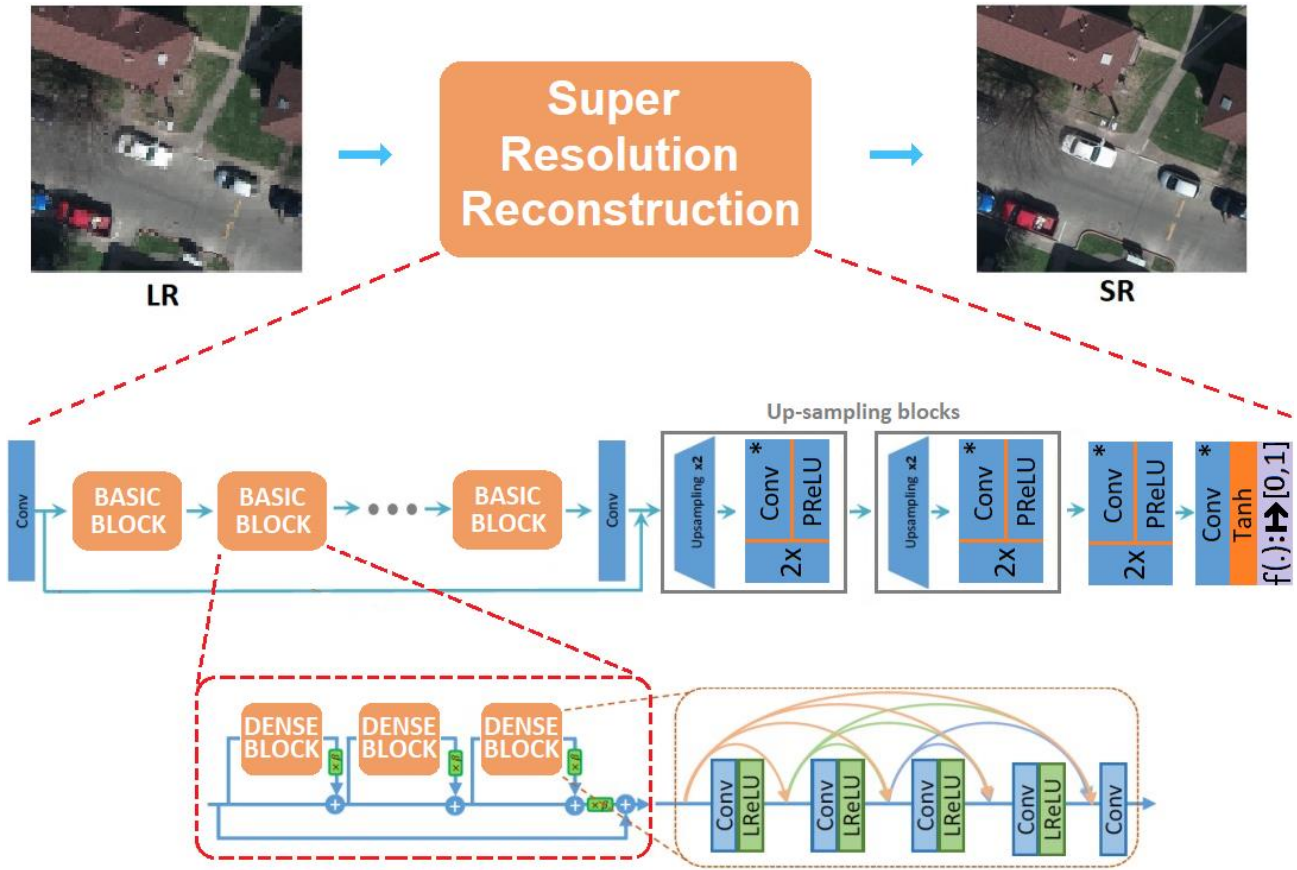

**Fig. 2.** The modified ESRGAN architecture is used in the proposed super-resolution model. Several RRDBs process the image while maintaining its low-resolution dimensions and upsampling is applied near the output of the model. We introduce two convolutional layers with PReLU activations between the upsampling units as well as two additional convolutional layers with PReLU activations after the last upsampling unit. Furthermore, we use a hyperbolic tangent activation function at the output and a rescaling layer just before the inferred SR image to adjust the values in the range [0,1]. Our modifications on the basic ESRGAN are shown with an asterisk on the right top area of the network components.

We propose this design choice after a series of observations made during trial-and-error experimentation with various content losses. Using MAE or MSE, the SRR model tends to predict tree and soil reconstructions with relatively low errors, but reconstructions are overly smooth and blurry. By not heavily penalizing errors of such depictions (trees, soil, or other complicated structures), we shift the burden of generating realistic reconstructions to the other losses (i.e., the nDSM and the adversarial loss). Accordingly, by greatly penalizing the reconstruction error of structures like houses and cars, the influence of the nDSM and the adversarial loss is reduced, and the model avoids the generation of high-frequency artifacts. Meyer [43] proposed an alternative probabilistic interpretation of the Huber loss which justifies its use on tasks dealing with aerial photography since these tasks generally contain significant noise and often have low quality: Huber loss minimization relates to minimizing an upper bound on the Kullback-Leibler divergence between the Laplacian distribution of noise in the ground-truth data and the Laplacian distributed prediction noise. Meyer further showed that the optimal transition point of the Huber function is closely related to the noise in the ground-truth data. Taking the above into account, we train the proposed models for minimizing the following combination of the three losses, as described above (content, nDSM, and adversarial loss):

$$L = \alpha L_{nDSM} + L_{content} + \beta L_{adversarial} \quad (4)$$

with $\alpha$ and $\beta$ being the weighting factors of the losses.

## IV. EXPERIMENTS AND RESULTS

We evaluate our methodology with two datasets, one containing images mainly of an urban area taken by an aircraft equipped with image and Ranging Laser Scanner (LiDAR) sensors and one dataset containing aerial images mainly of rural areas. Both datasets contain the corresponding DSMs and Digital Elevation Models (DEMs). This variety in landscapes enables us to assess the developed models' performance in different imagery dataset contexts and spatial resolutions. As mentioned before, the training of the models requires the nDSM of the area used in the training data while an nDSM is not required during inference. Still, this limitation only allows the use of datasets that include DSMs and DEMs such as the Data Fusion 2018 Contest Dataset (DFC2018) [44], [45] and the dataset containing the national LiDAR flyby of Luxembourg, conducted in 2019 by the country's administration for cadaster and topography [46]-[47]. The DFC2018 dataset is part of a set of community data provided by the IEEE Geoscience and Remote Sensing Society (GRSS). In this paper, we specifically use the Multispectral LiDAR Classification Challenge data. The RGB images of the DFC2018 dataset have a *5cm/pixel* spatial resolution while the LiDAR resolution is *50cm*/pixel. The data belongs to a *4172×1202 $m^2$* area and includes the University of Houston and its surroundings. The Luxembourg dataset contains RGB images of *20cm/pixel* spatial resolution, and the LiDAR resolution is *50cm/pixel* with a density of *15* points/ $m^2$. The Luxembourg dataset is in georeferenced raster format and uses the LUREF (EPSG 2169) coordinate system and projection. We use the datasets to train two models for the SRR × 4 task (one model for each dataset) and the DFC2018 dataset to train a model for the SRR × 8 task. We do not use the Luxembourg dataset to train a model on the SRR × 8 task because of the poor quality of the downscaled images.

*A. Training details*

We train our models with the Adam optimizer and a learning rate of *0.0001,* scaling the nDSM and the adversarial losses with factors $\alpha = 0.01$ and $\beta = 0.001$ respectively, as shown in Equation *4*. We also apply label smoothing of *0.2* to the GAN training and we pre-train the SRR models with MAE. This puts the weights in an appropriate configuration to avoid local minima and stabilize the GAN training [17]. The MAE of the pre-trained models also provides an indication of what constitutes a suitable value for the transition point of the Huber loss (content loss). Our experiments showed that a Huber loss transition point that is twice the MAE of the pre-trained models gives better results. In our experiments, we use × 4 and × 8 upsampling factors. For the × 4 upsampling experiments, we train the models with randomly cropped patches of size 520 × 520 pixels, which are downscaled with bicubic interpolation to LR inputs of size 130 × 130 (the models apply × 4 upsampling and thus the SR outputs match the size of the original HR images). For the × 8 upsampling experiment on the DFC2018 dataset, the 520× 520 random patches are downscaled via bicubic interpolation to LR inputs of size 65 × 65. The LR images in all experiments are created by downscaling the HR images via bicubic interpolation.

*B. Results*

Figures 3 and 4 show the results for the SRR × 4 models for the DFC2018 and the Luxembourg datasets respectively. Figure 5 shows the results of the SRR × 8 model for the DFC2018 dataset. The SRR × 4 model dealing with the DFC2018 dataset reconstructs finer image details and more high-frequency components in comparison to the model trained on the Luxembourg dataset. This is not surprising since the resolution of the images in the DFC2018 dataset is four times higher than the resolution of the images in the Luxembourg dataset and thus the model learns the data distribution of a more detailed scenery. This is also reflected in the performance metrics shown in Table I, i.e., the SRR model trained with the DFC2018 dataset achieves a higher Structural Similarity Index Measure (SSIM) [48] and a PSNR than the model trained with the Luxembourg dataset for the SRR × 4 task. The results also suggest that the effectiveness of the proposed SRR approach depends on the quality of the nDSM model used. This is one of the reasons why our approach works better on the DFC2018 dataset, as it has a more accurate nDSM. Since the quality of the nDSM model relates to the quality of the images contained in the dataset, the effectiveness of the proposed approach inherently relates to the quality of the images in the dataset. Hence, the results of the SSR × 4 model trained with the high-quality DFC2018 dataset are often indistinguishable from the ground truth HR images (a more detailed analysis of this is provided in Section IV.D regarding the limitations of the model). Table I also shows the values of PSNR and SSIM achieved by the ESRGAN trained on the datasets and the corresponding values achieved when bicubic interpolation is applied to the LR images. The overall findings indicate that the proposed model achieves higher scores on the PSNR and SSIM metrics than when bicubic interpolation or the ESRGAN model are applied. An analysis of the comparison between our model and the ESRGAN model is held in Section IV.C. We must note the difficulty in comparing the performance of our model with the performance of previous studies on SRR models applied to remote sensing imagery because the datasets involved in the comparison must contain DSMs.

TABLE I
PERFORMANCE OF THE PROPOSED APPROACH

|  |  | SSIM | PSNR (DB) |
|---|---|---|---|
| X 4 DFC2018 | **OUR** | **0.92** | **31.17** |
|  | ESRGAN | 0.91 | 30.43 |
|  | BICUBIC | 0.83 | 28.41 |
| X 4 LUXEMBOURG | **OUR** | **0.83** | **26.46** |
|  | ESRGAN | 0.81 | 26.1 |
|  | BICUBIC | 0.68 | 23.3 |
| X 8 DFC2018 | **OUR** | **0.88** | **28.62** |
|  | ESRGAN | 0.85 | 26.9 |
|  | BICUBIC | 0.66 | 24.55 |

The proposed SRR model performing × 4 upsampling recovers significant information content which was lost during the downscaling of the images (Figures 3-4). Various objects like cars and street poles are properly reconstructed and, in many cases, some fine details like shadows and street lines are almost identical to their HR counterparts. Buildings are also properly reconstructed with high-level details and large surfaces like rooftops are depicted with their original texture which was severely degraded during downscaling. The reconstructions with the least fidelity are those of trees and soil, which is something expected given the distribution complexity of their surfaces and their extreme diversity in visual representations. As expected, the results of the SRR model performing × 8 upsampling show lower fidelity because of the high information loss during the downsampling of the ground truth images. Regardless of the lower quality, the resulting images (Figure 5) recover a lot of details like difficult-to-identify train rails, rooftop textures, car shapes, road details, street poles and shadows. The significant image quality improvement observed at the output of the model in comparison to the LR input image reflects the quality metrics' improvement shown in Table I.

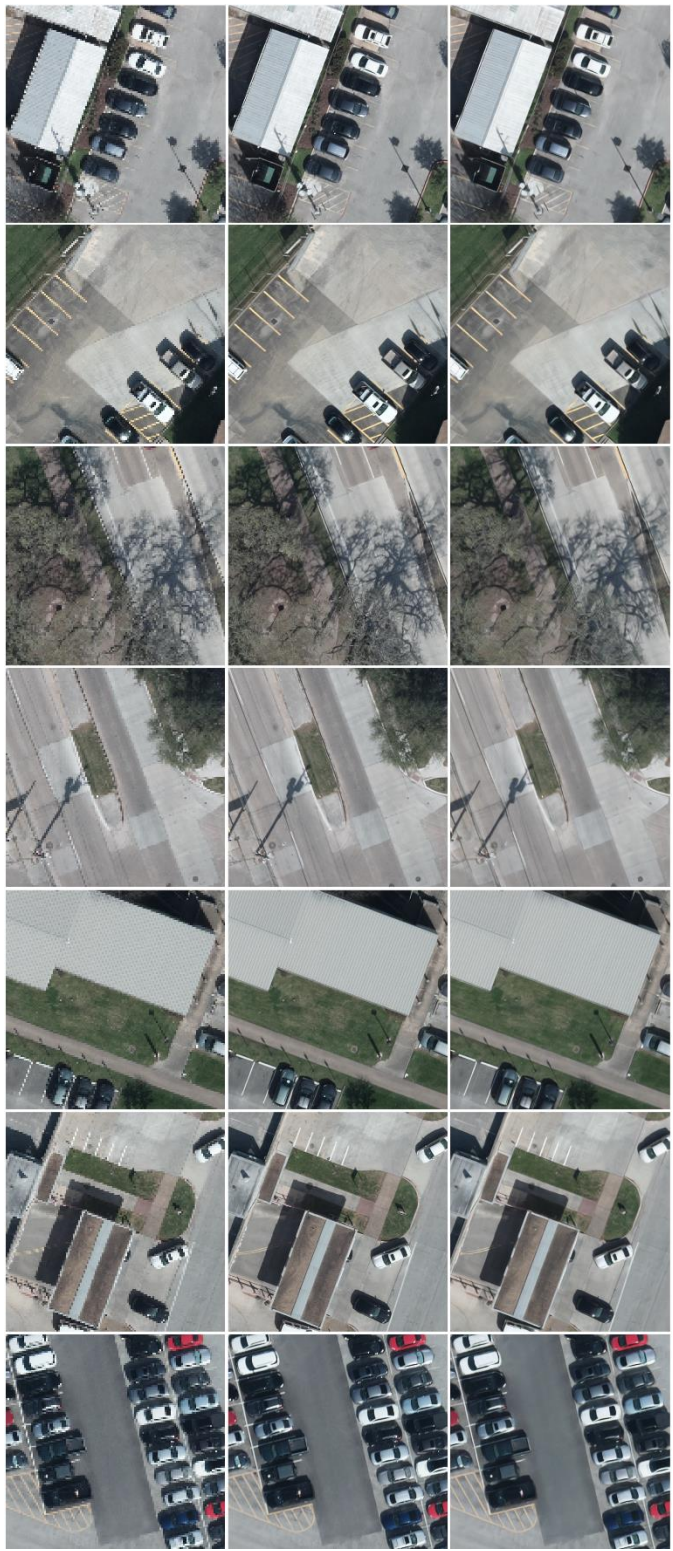

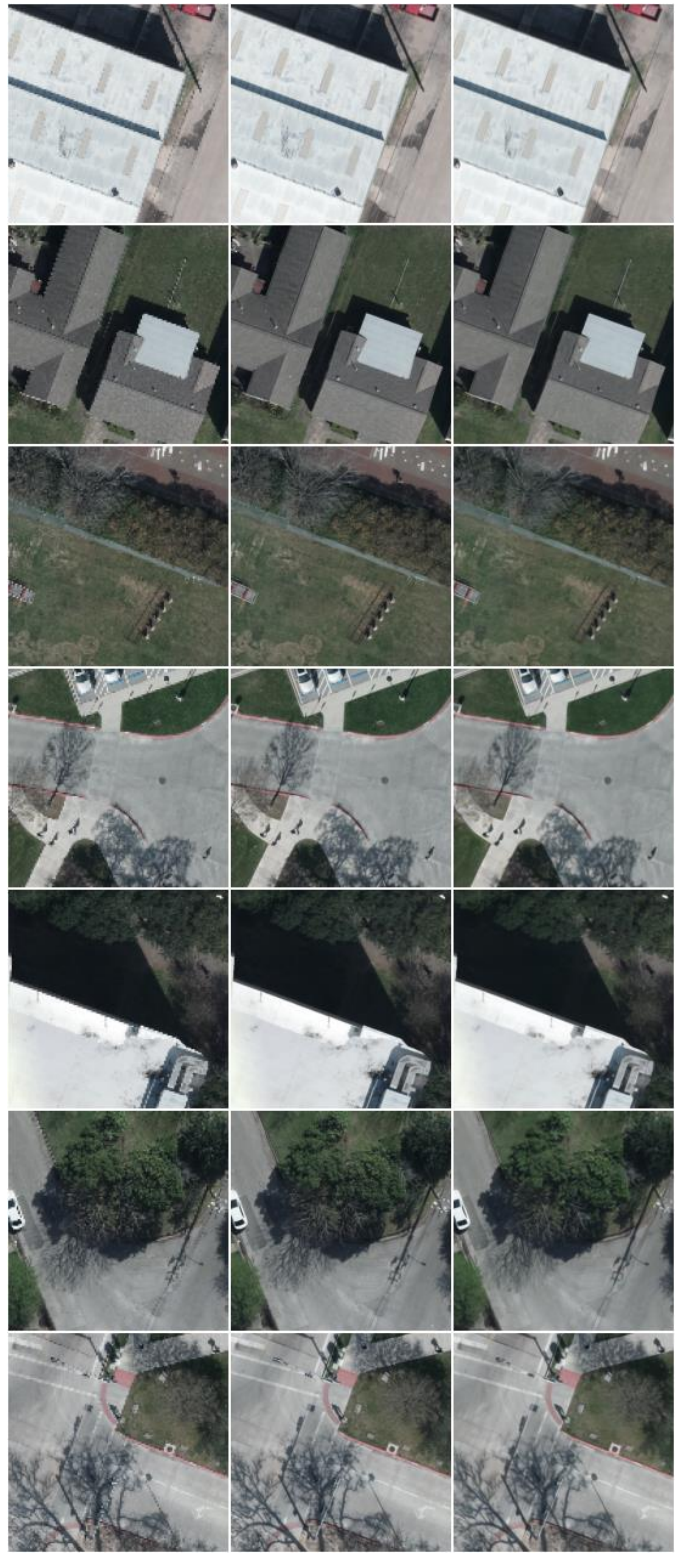

**Fig. 3.** SRR × 4 results for the DFC2018 dataset. Columns from left to right: LR, HR (ground truth), and SR images (inferred by our model).

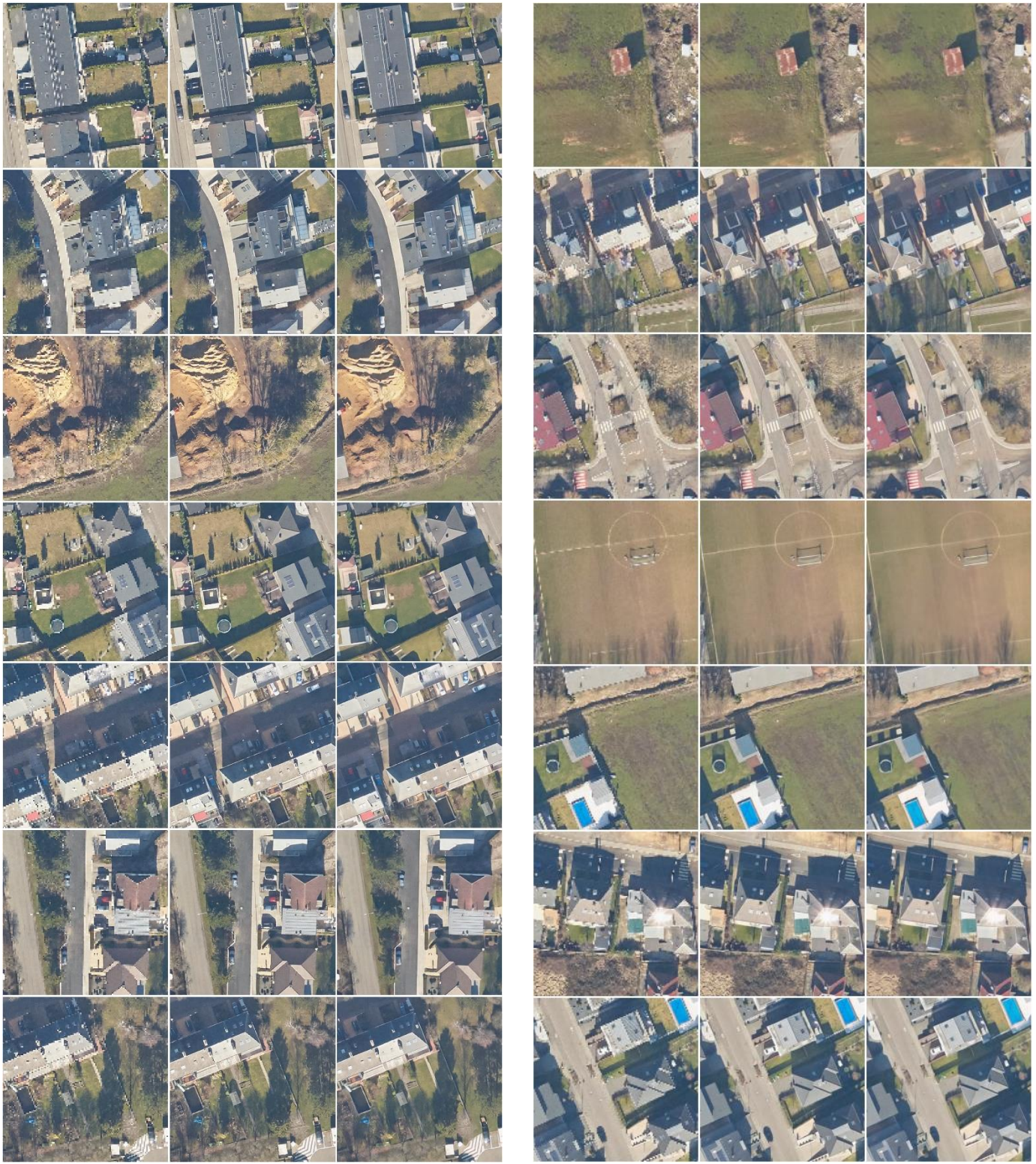

**Fig. 4.** SRR × 4 results for the Luxembourg dataset. Columns from left to right: LR, HR (ground truth), and SR images (inferred by our model).

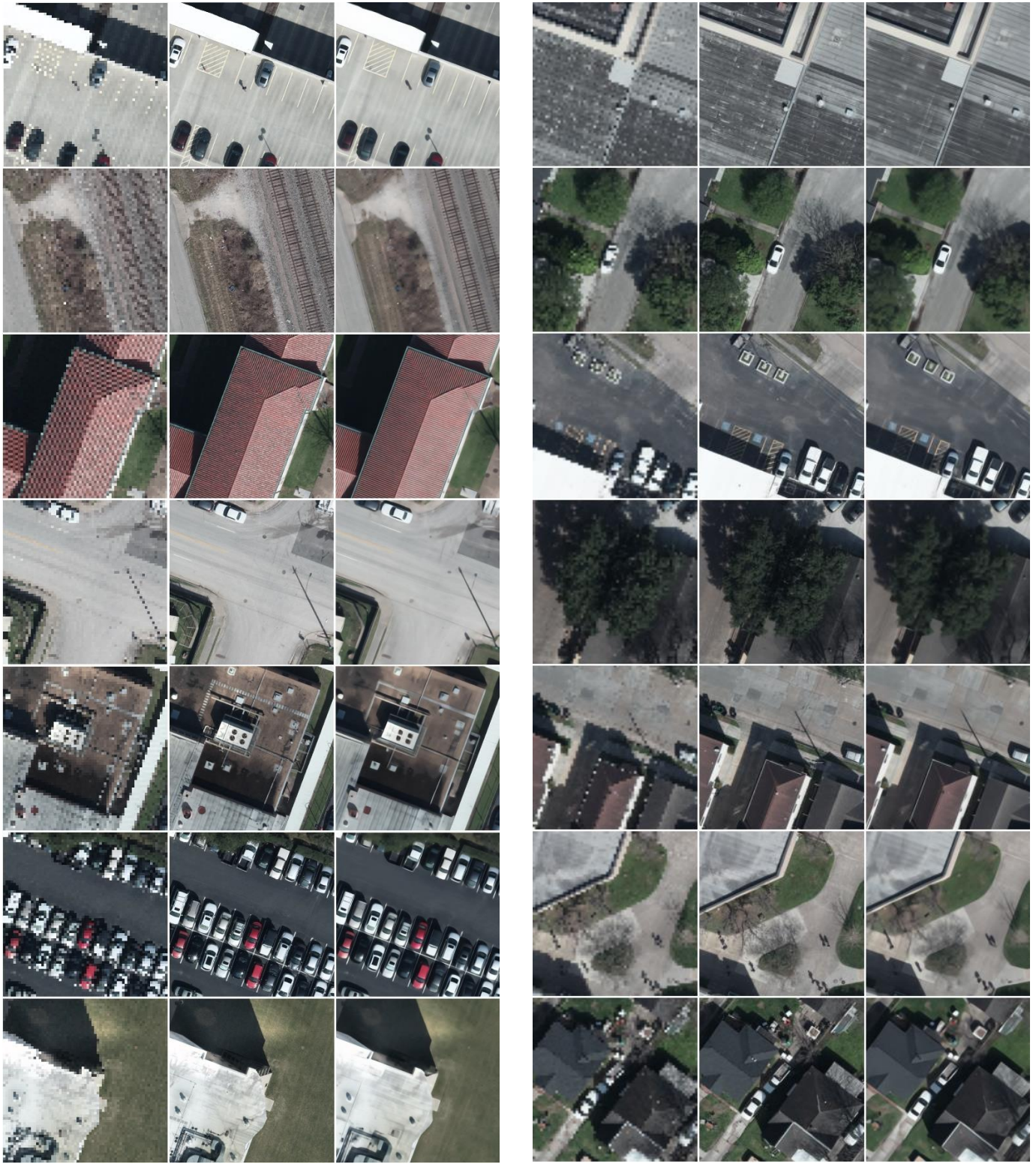

**Fig. 5.** SRR × 8 results for the DFC2018 dataset. Columns from left to right: LR, HR (ground truth), and SR images (inferred by our model).

The results shown in Figures 3-5 may encourage the use of the proposed model for upsampling LR images to feed other remote sensing models performing challenging tasks that may require inputs of a higher level of detail. Some examples of such tasks are humans counting, tree identification, car type identification, dumping detection, chimney detection, detection of parking spots for disabled people, power cable identification, etc. Such tasks are supported by our results as shown in Figures 3-5, i.e., the objects involved in these tasks are enhanced by our model and they become more evident (their detail level is enhanced).

*C. Comparison with the ESRGAN*

We give special importance to comparing our model with the ESRGAN model because the ESRGAN outperforms most SRR methods including the SRGAN (of which it is an improvement). The ESRGAN model is very popular and constitutes a suitable and established model for applying the SRR task to various applications. Furthermore, our model is developed based on similar concepts used by the ESRGAN (architecture, loss functions) while introducing important and necessary modifications to make it suitable for applying the SRR task on remote sensing imagery.

Despite being one of the best performing models for applying the SR task on natural images, ESRGAN's performance on aerial and remotely sensed imagery has a significant drawback: ESRGAN creates artifacts, especially on large textured surfaces that are very common in remotely sensed imagery[31]. We trained the ESRGAN model [49] on our datasets to compare its results with ours and assess the performance of our approach. We specifically used SSIM and PSNR as the comparison metrics. The resulting metrics' values are shown in Table I. Our approach achieves better performance on both metrics for both datasets and upsampling factors ($\times 4$ DFC2018, $\times 4$ Luxembourg and $\times 8$ DFC2018). Visual inspection of our approach and ESRGAN's results reveals no significant differences in the quality of the generated images except for some annoying artifacts created by the ESRGAN, especially on large flat surfaces. Some examples of these artifacts when applying the $\times 4$ SR task on remote sensing imagery with the ESRGAN are shown in Figure 6. Occasionally (on about 5% of the images produced by the ESRGAN), these artifacts are so acute that distort a portion of the SR image significantly. The images produced by the ESRGAN reveal its eagerness to reconstruct high-frequency components, a property that proves to be productive when working with natural images, but it is problematic when working with remotely sensed imagery. The main reason ESRGAN achieves lower scores on the PSNR and SSIM evaluation metric values than our model is the artifacts it produces. We confirmed this by calculating the evaluation metric values only for the images produced by ESRGAN that do not contain artifacts (the screening was conducted with visual examination). The metric values scored by the ESRGAN's results after excluding the images containing artifacts were closer to the scores achieved by our method than when including the distorted images. The generated artifacts are even more evident when we use the ESRGAN for the $\times 8$ SR task. Figure 7 shows some comparative examples of using the ESRGAN and our method for the $\times 8$ SR task.

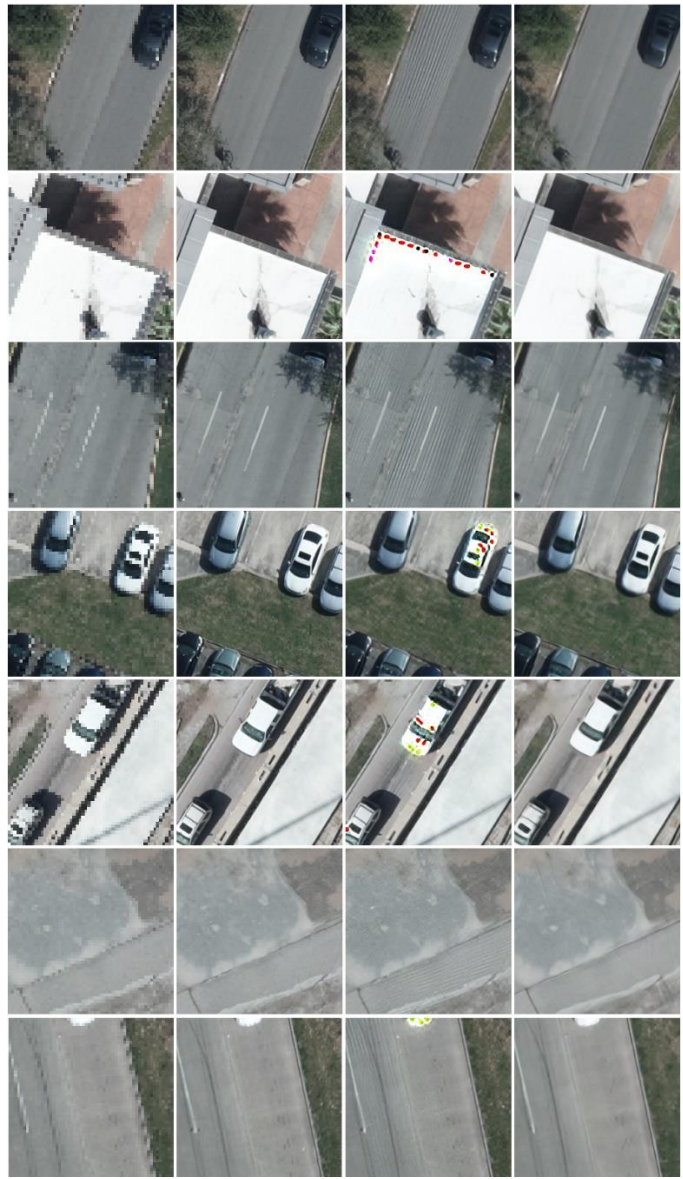

**Fig. 6.** Examples of the artifacts created by the ESRGAN on large surfaces depicted in remotely sensed imagery for $\times 4$ SR. From left to right: the LR images, the original HR images, the SR images generated by the ESRGAN and the SR images generated with our approach. Besides its occasional difficulty to reconstruct some objects (e.g., cars), ESRGAN adds artifacts on large surfaces (e.g., roads and roofs). On the contrary, our model does not.

Figure 7 shows that there are more artifacts (both in number and intensity) and more distortion on large surfaces like building roofs and roads compared to the artifacts observed in the application of the $\times 4$ SR task with the ESRGAN as shown in Figure 6.

We believe that ESRGAN produces these artifacts because of the nature of its perceptual loss: the pre-trained VGG19 model that is used for obtaining the perceptual loss of the inferred SR images is not a good option for remotely sensed imagery. VGG19 is pretrained on the ImageNet dataset that has a very different data distribution than remotely sensed imagery.

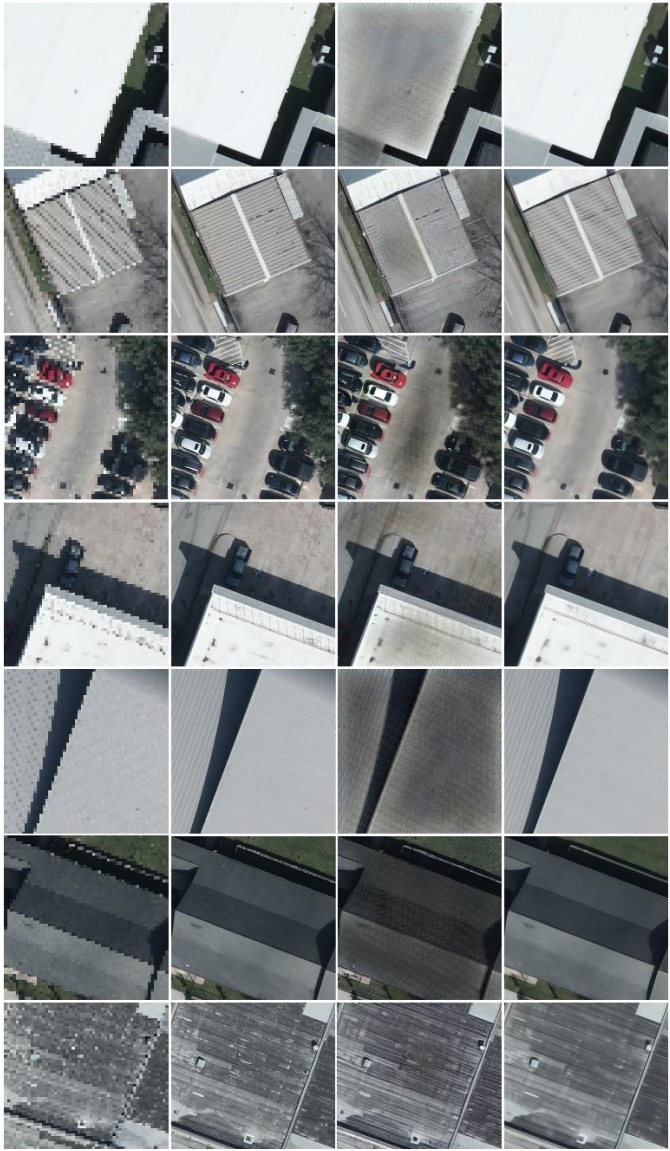

**Fig. 7.** Examples of the artifacts created by the ESRGAN on large surfaces depicted in remotely sensed imagery for $\times 8$ SR. From left to right: the LR images, the original HR images, the SR images generated by the ESRGAN and the SR images generated with our approach. The ESRGAN tends to add a significant amount of texture and lines to the roads and building roofs.

This difference in data distribution renders the feature mapping layer used for the perceptual loss incapable of calculating appropriate remotely sensed image representations. Therefore, the essence of what constitutes a high-quality remotely sensed image cannot be captured by the perceptual loss calculated with a model pre-trained on a different data distribution e.g., the VGG19 used by the ESRGAN model. Our approach proposes a solution to this problem by replacing the ImageNet classifier with a pre-trained nDSM prediction model. The nDSM model is trained on predicting the height maps of remotely sensed images and thus it is domain-specific and relevant to the images used in the SR task.

*D. Limitations*

The main limitations of our work have to do with the upsampling factor targeted, e.g., $\times 4$, $\times 8$, etc., the spatial resolution of the HR images (training images) and the proper alignment between the HR images and their DSMs. We tested our model on up to a $\times 8$ upsampling factor and the results are decent (Figure 5). We do not show results from experiments with larger upssampling factors because beyond $\times 8$ upsampling, the SR task becomes overwhelmingly ill-posed [29], [37], [38]: it is extremely difficult for LR images (downscaled images) to contain information regarding fine details of the original HR images. Similarly, SRGAN and ESRGAN focus solely on a $\times 4$ resolution upsampling factor [17], [28]. In contrast to the case of natural images, the benefits of applying the SR task on remotely sensed imagery for $\times 8$ upsampling are de facto more profound because of the high cost of high-resolution imagery. This makes our approach more suitable for remotely sensed imagery since SRGAN and ESRGAN are optimized for $\times 4$ spatial resolution upsampling. However, the experiments we conducted revealed that our model does not perform well for upsampling factors greater than $\times 8$.

We also experimented with the spatial resolution of the HR images used for training the model. We used a dataset available from the national Open Government Data (OGD) initiative of Austria [50]. The dataset contains RGB images of *40cm/pixel* spatial resolution, and the LiDAR resolution is *40cm/pixel* (the acquired RGB images were of *20cm/pixel* resolution, but the publicly available data is downscaled to *40cm/pixel* resolution). At this lower spatial resolution, the nDSM model that provides the perceptual loss does not generalize well, probably because it loses its ability to detect low-level features like corners, lines, shadow edges, or other features that allow it to predict the heights of the objects in the image. Thus, the advantages of using an nDSM-based perceptual loss are lost which explains why our model operating on images of lower spatial resolution performs similarly to the case of using a combination of a MAE and a GAN loss alone, i.e., without the nDSM-based perceptual loss. In other words, while the nDSM-based perceptual loss has a significant contribution to the performance of the model when the spatial resolution of the HR images is adequate for the nDSM model to learn how to infer the heightmaps of the images, this contribution ceases to exist when the resolution of the HR images (and the quality of the data) prevent the nDSM model from generalizing well.

Another limitation of our model related to the nDSMs of the HR images is the proper alignment between them. For example, the nDSMs of the DFC2018 dataset are properly aligned with the RGB images and this allows the model to achieve very good results for both upsampling factors ($\times 4$ and $\times 8$). The nDSMs of the Luxembourg dataset are not perfectly aligned and we suspect that this plays a significant role in the model's inability to perform well on the $\times 8$ SR task for the specific dataset (the results are not better than using a MAE and a GAN loss only). The nDSM model trained on the Luxembourg dataset is performing much worse than the nDSM trained for the DFC2018 dataset. Specifically, the nDSM model for the DFC2018 dataset has a MAE of *0.54m* on its test set while the nDSM for the Luxembourg dataset has a MAE of *0.83m*. This great difference is attributed partly to the lower resolution of the Luxembourg dataset and mostly to the misalignment between its nDSMs and RGB images. To investigate the effects of the misalignment problem, we trained the model on the DFC2018 dataset with the nDSMs shifted by a constant value of 2 pixels in random directions (up, down, left, right) for each image and the nDSMs distorted with a random transformation (small rotation, slight skew). In the case of the constant shifting of the nDSMs, the results were very similar to the unaltered nDSMs. On the contrary, in the case of random small transformations, we observed a significant reduction in the performance metrics (in the range of 5-10%). This suggests that the quality of the nDSMs is also important for our method to achieve its potential.

## V. Conclusion

This paper proposes a super-resolution (SR) reconstruction (SRR) model that works with remotely sensed images, addressing the limitations of existing state-of-the-art models by including auxiliary information that is important during the model's training phase, i.e., including the corresponding normalized Digital Surface Model (nDSM) of the remote sensing imagery dataset. In other words, the proposed SRR model, during training, considers the difference between the nDSM inferred by the calculated SR image and the ground truth high resolution (HR) image, instead of using a perceptual loss. Furthermore, the contribution of this paper includes applying some architectural changes to the ESRGAN model and employing a Huber loss as a content loss to mitigate the difficulties imposed by remotely sensed images. Visual inspection, together with the significant improvement of the SSIM and PSNR metrics of the inferred SR images obtained, suggest that the proposed model is suitable for the SRR task and can cope with popular and notorious remote sensing imagery limitations such as big surface textures and lower spatial resolution. Summing up, this paper shows that an nDSM-based loss seems to be suitable for the SRR task on remote sensing imagery, supplying the model with enriched pixel-level information. This approach allows us to detect objects of interest that were otherwise impossible to identify such as car types, powerlines, parking spots, chimneys and tree types. This information could be very useful to city planners, policymakers, operators of municipalities and local communities.

**Funding:** This project received funding from the European Union's Horizon 2020 Research and Innovation Programme under Grant Agreement No. 739578 and the Government of the Republic of Cyprus through the Deputy Ministry of Research, Innovation and Digital Policy.